\documentclass[letterpaper, 10 pt, conference]{ieeeconf}  

\IEEEoverridecommandlockouts                              

\newcommand{\RNum}[1]{\uppercase\expandafter{\romannumeral #1\relax}}

\usepackage{graphics} 
\usepackage{graphicx}
\graphicspath{{./figure/}}
\usepackage{amsfonts}%
\usepackage{bm}
\usepackage[noadjust]{cite}
\usepackage{hyperref}
\usepackage{amsmath} 

\usepackage{amssymb}  
\usepackage{algorithm}
\usepackage{algorithmic}
\usepackage{booktabs}
\usepackage{multirow}
\usepackage{cleveref}
\usepackage{color}
\usepackage{array}
\newcolumntype{C}[1]{>{\centering\arraybackslash}m{#1}}
\newcolumntype{N}{@{}m{0pt}@{}}
\usepackage{upgreek}
\usepackage{tabu}
\usepackage{siunitx}
\usepackage{nccmath}
\begin{document}






\title{DelTact: A Vision-based Tactile Sensor Using Dense Color Pattern}

\author{Guanlan~Zhang,~Yipai~Du,~Hongyu~Yu~and~Michael~Yu~Wang,~\IEEEmembership{Fellow,~IEEE}
\thanks{G. Zhang is with the Hong Kong University of Science and Technology (Guangzhou), email: gzhangaq@ust.hk. Y. Du, H. Yu and M. Y. Wang are with the Hong Kong University of Science and Technology, Hong Kong, e-mail: (yduaz, hongyuyu, mywang)@ust.hk. M. Y. Wang is also with the HKUST Shenzhen-Hong Kong Collaborative Innovation Research Institute, Futian, Shenzhen.}
\thanks{This work was supported by the Hong Kong Innovation and Technology Fund (ITF) under Grant ITS/104/19FP, and supported in part by the Project of Hetao Shenzhen-Hong Kong Science and Technology Innovation Cooperation Zone (HZQB-KCZYB-2020083).}}

\maketitle
\begin{abstract}

Tactile sensing is an essential perception for robots to complete dexterous tasks. As a promising tactile sensing technique, vision-based tactile sensors have been developed to improve robot performance in manipulation and grasping. Here we propose a new design of a vision-based tactile sensor, DelTact. The sensor uses a modular hardware architecture for compactness whilst maintaining a contact measurement of full resolution ($798\times586$) and large area ($675$mm$^2$). Moreover, it adopts an improved dense random color pattern based on the previous version to achieve high accuracy of contact deformation tracking. In particular, we optimize the color pattern generation process and select the appropriate pattern for coordinating with a dense optical flow algorithm under a real-world experimental sensory setting. The optical flow obtained from the raw image is processed to determine shape and force distribution on the contact surface. We also demonstrate the method to extract contact shape and force distribution from the raw images. Experimental results demonstrate that the sensor is capable of providing tactile measurements with low error and high frequency (40Hz).

\end{abstract}

\section{Introduction}
\label{sec: Intro}

Through millennia of evolutionary processes, humans have evolved the sense of touch as a critical sensory method to perceive the world. Delicate tasks, including tactile perception, grasping different shaped objects, and manipulation of tools, can be completed with fluency and insight given direct sensory feedback. As the ratio between intelligent robots and humans has been increasing worldwide, robotic systems are pursuing more dexterity in the face of contact-rich scenarios, where tactile sensors are being developed to detect the necessary tactile information for robot interaction with objects and environments. 


Conventional tactile sensors utilize transduction materials such as piezoresistive, capacitive, and piezoelectric components to convert physical contact into digital signals for processing at high speed. But sensitivities to environmental temperature, vibration, and electrical interference are issues to be solved. Also, difficulties exist in data acquisition connection and interpretation software which further inhibits the broader application of such sensors \cite{zou2017novel}. 

In recent years, research into vision-based tactile sensors has been growing due to superiorities in low cost, easy fabrication, high durability, and multi-axial measurements. 
Developments in digital cameras have made capturing contact situations with high-quality images of low cost and easy to interface with. Moreover, progress in computer vision and deep learning assists in transferring knowledge from visual perception to tactile perception, enabling faster analysis of high dimensional tactile representation in larger images. 

\begin{figure}
    \centering
    \vspace{2mm}
    \includegraphics[width=0.47\textwidth]{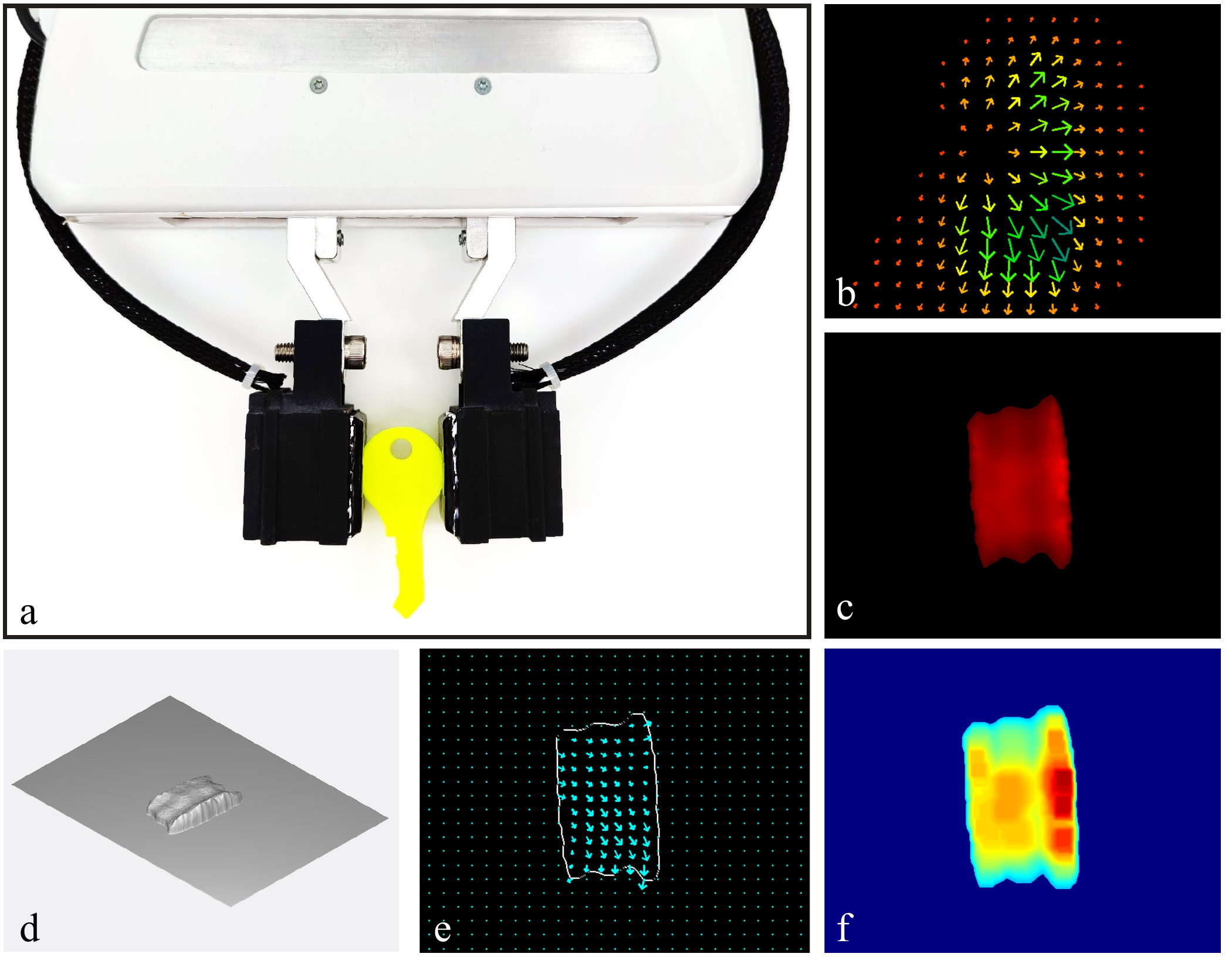}
    \vspace{-0.4cm}
    \caption{a) Two DelTact sensors are mounted on a FE Gripper of Panda Robotic Arm to grasp a yellow key. b) Visualized optical flow. c) Gaussian density plotted in hot map. d) Isometric view of the shape reconstruction. e) Estimated shear force distribution. f) Estimated normal force distribution.}
    \label{fig: cover}
    \vspace{-0.7cm}
\end{figure}

Representative studies including GelSight \cite{yuan2017gelsight}, GelSlim \cite{taylor2021gelslim3}  and Digit \cite{lambeta2020digit} demonstrated the advance of vision-based tactile sensor in contact measurement. These sensors adopted photometric stereo technology to obtain a precise depth estimation. And, similar to GelForce \cite{vlack2004gelforce}, they measured the surface deformation by tracking dot markers. However, to guarantee the accuracy of photometric stereo, the dot tracking methods cannot achieve full resolution. The deformation in the undetected regions relied on interpolation from the near measurements, which might cause information to lose. To obtain a full-resolution surface deformation tracking, Sferrazza et al. \cite{sferrazza2019design} and Kuppuswamy et al. \cite{kuppuswamy2020soft} captured the movement of pattern/particles with higher density on the contact substrate. But the quality of the tracking remained for further improvement and the results were utilized solely for force \cite{sferrazza2019design} or depth \cite{kuppuswamy2020soft}. Therefore, our motivation is to develop a sensor that can first, achieve full resolution measurement of the surface deformation and then, extract more contact information including depth and force.

In this paper, we present DelTact, a new version of the vision-based tactile sensor based on our previous framework \cite{du2021high}. This name comes from the abbreviation of the sensor's main feature: using a dense color pattern to capture tactile information. The sensor is designed to be compact and convenient in integrating itself into modern robotic systems such as grippers and robot fingers with online sensing of shape and force at high spatial and temporal resolution. Our work thus makes three main contributions to this field:

\begin{itemize}
	\item Presenting a new modular hardware design of a vision-based tactile sensor. This sensor has a simple structure, large sensing area ($675$mm$^2$), while its size is maintained compact (shown in Table. \ref{Tab:sensor comp}).
 	\item Proposing a parametric optimization framework of the previous random color pattern \cite{du2021high} with indentation experiment to track 2D displacement field with an accuracy of submillimeter scale at full resolution.
	\item Integrating online tactile measurement algorithms into software to extract contact shape and force distribution from the optical flow at a high spatial ($798\times586$) and temporal resolution (40Hz).
\end{itemize}


The paper proceeds as follows: Section. \ref{sec: related works} introduces related works on vision-based tactile sensor designs and information processing methods. In, Section. \ref{sec: hardware}, we give a complete description of design and fabrication of the proposed sensor. In Section. \ref{sec: software}, algorithms including raw image preprocessing, surface deformation measurement, and contact information extraction are presented. In Section. \ref{sec: experiment}, pattern selection and tactile measurement experiment results are shown and analyzed. Finally, in Section. \ref{sec: conclusion}, discussion and conclusion with future research are identified.


\section{Related Work}
\label{sec: related works}

\subsection{Vision-based Tactile Sensor}
Early in 2001, Kamiyama et al. developed a vision-based tactile sensor \cite{kamiyama2001vision}, where colored markers were deployed in a transparent elastomer and tracked by a CCD camera to measure the gel deformation at different depths. The concept of this prototype later was improved into GelForce \cite{vlack2004gelforce}, which could obtain complete information on contact force (i.e., direction, magnitude and distribution). 

Continued research based on the GelForce-type working principle focused on a more compact form factor hardware with broadened functionalities of the sensor to better integrate with robotic systems such as robot hands and grippers. Yamaguchi et al. \cite{yamaguchi2016combining} proposed a FingerVision sensor to combine visual and tactile sensing with only one monocular camera. Lepora et al. \cite{ward2018tactip} introduced the TacTip family with a bio-inspired data acquisition system to simulate mechanoreceptors under human skin and detect contact information. Sferrazza et al. \cite{sferrazza2019design} presented a high-resolution tactile sensor with randomly distributed fluorescent markers and used optical flow tracking to achieve high-accuracy force sensing. Kuppuswamy et al. \cite{kuppuswamy2020soft} showed a Soft-bubble gripper with a pseudorandom dot pattern to estimate shear deformation. These sensors are all characterized by simple structures and easy fabrication.

Another series of GelSight-type sensors adopted a retrographic sensing technique to obtain high-resolution 3D deformation. Works regarding GelSight were demonstrated by Yuan et al. \cite{yuan2017gelsight} and Dong et al. \cite{dong2017improved}, who cast colored light onto a Lambertian reflectance skin to measure the surface normal of deformation directly and reconstructed the dense accurate 3D shape. To further reduce the size of the sensor for convenient installation onto grippers, GelSlim 3.0 \cite{taylor2021gelslim3} was developed with optimized optical and hardware design. Padmanabha et al. \cite{padmanabha2020omnitact} showed a finger-size touch sensor, OmniTact, to perform multi-directional tactile sensing. Lambeta et al. \cite{lambeta2020digit} released their fingertip Digit sensor with integrated circuit design at low cost for extensive application in robot manipulation. 
In our work, we aim to design our sensor that combines the advantages of these two types, that is to achieve full resolution multi-modality contact perception and have a simpler structure and less restricted optical requirement than the Gelsight-type sensor.

\subsection{Tactile Information Extraction}

The origin signal received by the vision-based tactile sensor contains diverse compound information, which is dependent on the contact condition between sensor and environment. Furthermore, to achieve dexterity in challenging tactile-related tasks such as tactile exploration, grasping, manipulation, and locomotion,  tactile information extraction algorithms generally have capacity in multi-modality measuring of contact and versatility in recognizing various levels of features with models \cite{li2020review}.

Existing tactile sensors directly obtained low-level features such as deformation \cite{yamaguchi2016combining}, texture \cite{dong2017improved}, contact area localization \cite{donlon2018gelslim}, geometry reconstruction \cite{yuan2017gelsight} and force estimation \cite{yuan2017gelsight,sferrazza2019design} at the contact site.
Algorithms with low complexity could solve the problem using linear regression \cite{dong2017improved}, principal component analysis (PCA) \cite{she2019cable}, and graphic features such as entropy \cite{yuan2015measurement}, Voronoi feature \cite{cramphorn2018voronoi} and Gaussian density \cite{du2021high}.
Besides, complicated tasks that require high-level information have been performed, including object recognition \cite{yuan2017connecting}, localization of dynamic objects \cite{li2014localization}, simultaneous localization and mapping on objects \cite{bauza2019tactile}, and slip detection \cite{dong2017improved}. 
Learning-based methods may be preferred in such tasks to analyze high-dimensional tactile images with good generalization and accuracy.
In our work, we aim at using cost-effective algorithms to estimate low-level contact information as a proof of concept for our tactile sensing method.


\section{Hardware Design} \label{sec: hardware}

For the hardware part, three principles are suggested as guidelines in the sensor design.

\begin{enumerate}

    \item \textbf{Robustness}: The sensor should provide accurate and stable performance. This requires higher mechanical strength for longer service life and fewer noises during image capturing. 
    \item \textbf{Compactness}: With a compact size, the sensor is enabled for better integration with robot fingers to perform manipulation tasks under different scenarios, especially in a narrow space.
    \item \textbf{Easy to Use}: The sensor is easy to install, operate and maintain. Electrical parts including signal and power wires are also convenient for connection.

\end{enumerate}

\begin{figure} [ht]
    \centering
    \includegraphics[width=0.48\textwidth]{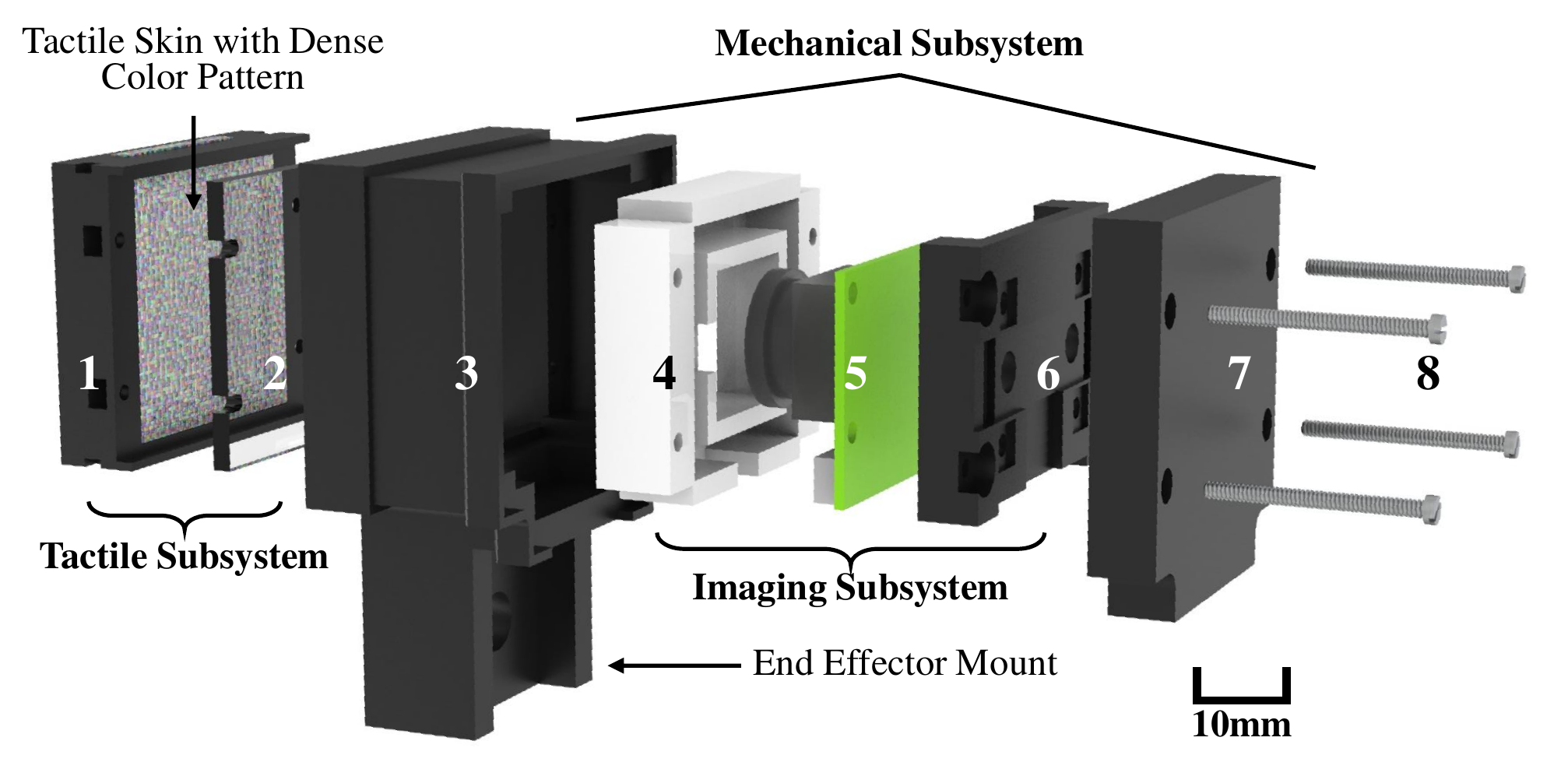}
    \vspace*{-3mm}
    \caption{Mechanical configuration of DelTact (explosive view). The parts are 1. tactile skin base; 2. acrylic plate; 3. sensor shell; 4. light holder; 5. camera; 6. camera holder; 7. sensor cover; 8. screws.}
    \label{fig:explod sensor}
    \vspace{-0.5cm}
\end{figure}

\subsection{System Configuration} \label{subsec: System Configuration}


Based on the design principles, the system configuration of DelTact is elaborated below. It consists of three subsystems (i.e., tactile subsystem, imaging subsystem, and mechanical subsystem) underlying the primary function of the sensor. The subsystems can be further disassembled into eight individual parts. Each part is designed for the least space required to achieve as much compactness as possible. Details about the subsystems are presented as follows.

\subsubsection{Tactile Subsystem}

The tactile subsystem comprises a tactile skin base and an acrylic plate. The tactile skin base (part 1 in Fig. \ref{fig:explod sensor}) is a black housing frame that fixes the tactile skin with dense color pattern at the bottom and avoids skin detachment from the sensor. The generation of dense color pattern is presented in Section \ref{subsubsec: pattern generate}.

For the tactile skin material, we choose a transparent soft silicone rubber (Solaris™ from Smooth-On, Inc.). The Solaris™ satisfies the requirement for both softness and toughness with a shore hardness of 15A and tensile strength of 180 psi. The thickness of the tactile skin is 12 millimeters and a surface area of $36$mm$\times34$mm is obtained with fillets on edges to reduce wear. To guarantee enough support against excessive deformation under external load, a 2-mm thick rectangle acrylic plate (part 2 in Fig. \ref{fig:explod sensor}) is attached tightly to the back of the skin.

\begin{figure} 
    \centering
    \vspace*{2mm}
    \includegraphics[width=0.4\textwidth]{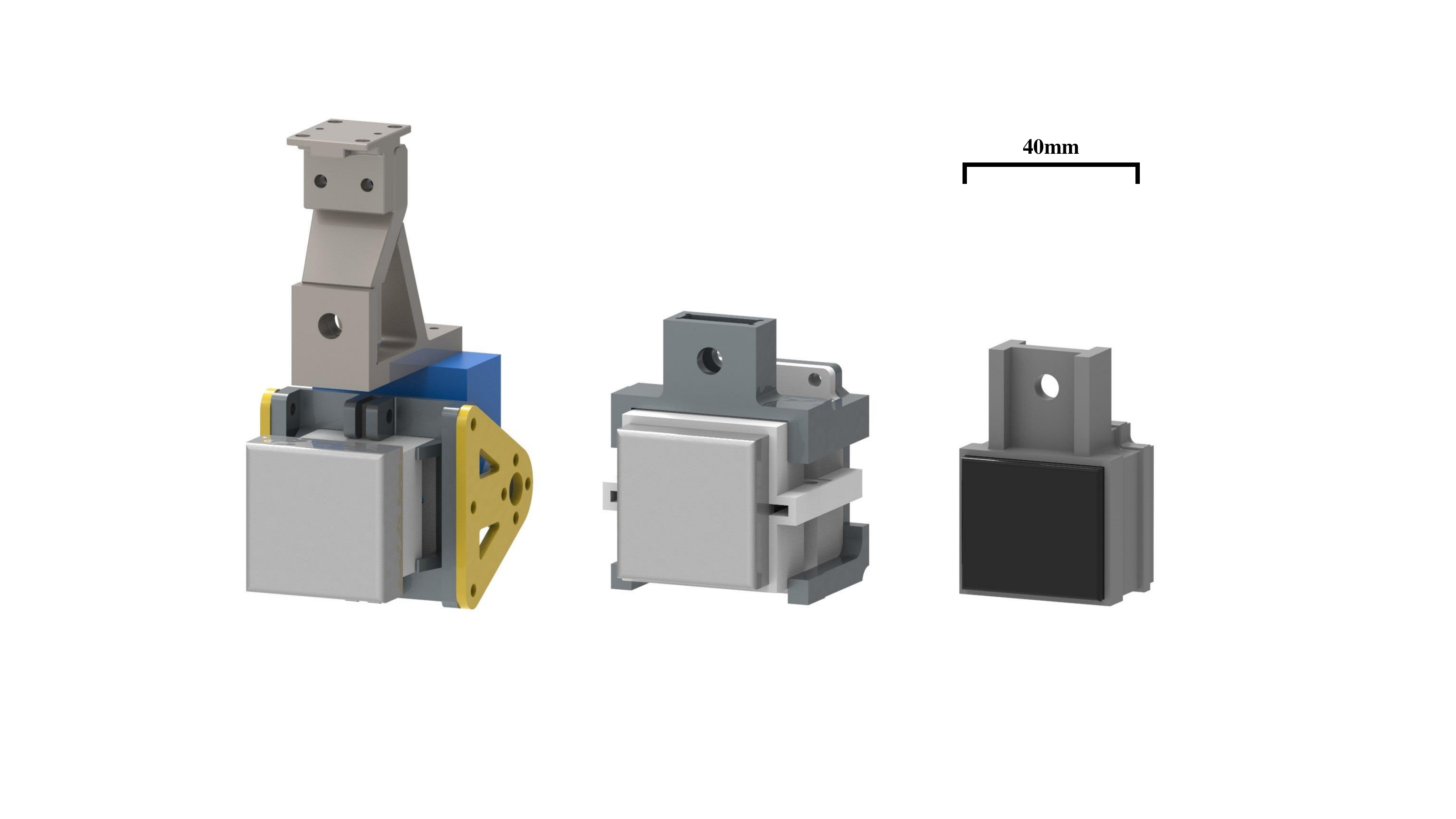}
    \vspace*{-1mm}
    \caption{3D model of previous Gecko-enhanced tactile sensor (left, $52 \times 108 \times 69$ mm$^3$), FingerVision sensor (middle, $51 \times 65 \times 52$ mm$^3$) and DelTact (right, $39 \times 60 \times 30$ mm$^3$).}
    \label{fig:sensor comp}
    \vspace{-0.5cm}
\end{figure}

\subsubsection{Imaging Subsystem}

The imaging subsystem consists of a light holder and a camera module (part 4 and 5 in Fig. \ref{fig:explod sensor}). The light holder is made of semitransparent white resin, and a light strip with five 5050 SMD LEDs is inserted into the holder. The strip is connected with a resistor of $750\Omega$ in series and powered by a $5$V DC power source to reach the desired illuminance. Owing to the scattering inside the light holder, light from LEDs is diffused to reduce overexposure. 

For the camera, the Waveshare IMX219 Camera Module with a short fisheye lens is chosen to achieve a 200-degree FOV and close minimum photographic distance. This camera is consistent with our compact design principle and able to acquire images with a high resolution of $1280\times720$ at 60 frames per second. Regarding signal transmission, the camera is connected to an Nvidia Jetson Nano B01 board, where the image can be directly processed with CUDA on board or sent to another PC with ethernet. To integrate the camera into the system, a camera holder (part 6 in Fig. \ref{fig:explod sensor}) is inserted to lock the camera with two M2 screws, which also fixes and stabilizes the camera and protects the camera circuit. 

\subsubsection{Mechanical Subsystem}

The mechanical subsystem includes the sensor shell and the sensor cover (part 3 and 7 in Fig. \ref{fig:explod sensor}). The purpose of designing the shell and the cover is to encapsulate the tactile sensing parts from outside interference and achieve maximum compact assembly. Therefore, the shell and cover are opaque and fully enclose the sensor to block external light and dust. The wall thickness of these components is 1.5 millimeters to ensure sufficient strength. Concerning flexible assembly and reducing relative slip, the cover has four snap-fit cylinders connected to the camera holder.

All the sensor components are assembled by four 19-mm long M1.6 screws, and each part can be maintained or replaced within minutes thanks to the modular design. A 20-mm long end-effector mount (shown in Fig. \ref{fig:explod sensor}) is set on the shell to work with the external connections. The geometry and position of the mount can be easily redesigned to fit into different grippers. The overall dimension of the sensor is $39\times60\times30$ mm$^3$ with the end-effector mount counted in. As shown in Fig. \ref{fig:sensor comp}, the size of DelTact is significantly reduced compared to the previous sensors \cite{du2021high}\cite{pang2021viko}, while the sensing area is almost maintained the same. 

\subsection{Fabrication Process} \label{subsec: Fabrication Process}

Fabrication of the sensor takes several steps into account.
To begin with, for preparation, all the mechanical components and a mold for silicone casting are fabricated using a photocuring 3D printer with black or white tough epoxy resin from Formlabs. This guarantees higher accuracy, stronger mechanism and smoother surface. Then the solvents of two-part Solaris™ silicone are mixed in a $1:1$ ratio and rest in a vacuum pump to remove air bubbles. The mixture is poured gently into the mold to cure in desired shape. This mold can cast four modules at a time. Meanwhile, the tactile sensing base is put into the mold to bind together with the tactile skin. 

The acrylic plate is laser cut and filmed uniformly with a layer of prime coat (DOWSIL™ PR-1200 RTV from Dow, Inc.) to enhance bonding between the silicone. It is also put onto a tactile sensing base whilst the silicone is curing. The mixed gel cures in 16 hours at room temperature ($23^\circ$C), and a heating process at $65^\circ$C in a constant temperature cabinet can effectively reduce this time to 8-10 hours. When formed, an elastomer layer adheres firmly under the acrylic plate and serves as the deformation interfacing substrate. 

\begin{figure}
    \centering
    \vspace*{2mm}
    \includegraphics[width=0.43\textwidth]{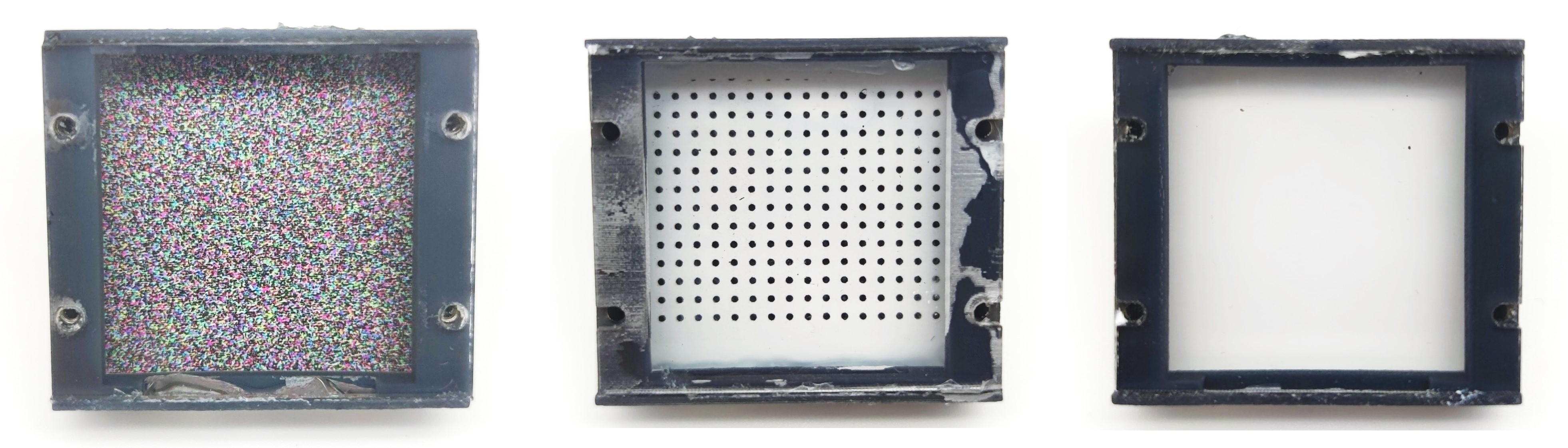}
    \vspace*{-1mm}
    \caption{Tactile sensing skin with three types of pattern (from left to right: dense color pattern, dot matrix, transparent).}
    \label{fig:tactile base}
    \vspace{-7mm}
\end{figure}

The dense color pattern is applied onto the transparent silicone surface with water transfer printing technique. The paint of the pattern is printed on a dry ductile soft film sticker. When contacted with water, the film becomes adhesive and adheres to the silicone surface. After the moisture is evaporate, the color pattern will remain tightly on the silicone. Different types of pattern stickers can be made to switch between different modalities, e.g., dense flow, sparse dots and transparent gel (shown in Fig. \ref{fig:tactile base}). When the sticker is dried, two thin layers of protection silicone (Dragon Skin™ 10 FAST from Smooth-On, Inc.) are coated on the pattern. White pigment is added to the inner layer to enhance imaging brightness and disperse light from the LEDs. Compared to the past design of spraying a frosted paint layer \cite{pang2021viko}, this method is simpler but more durable and takes less time to achieve the same effect. The outermost layer is sealed with black pigment to isolate potential external light disturbance and block background interference.

Finally, all eight components are assembled. Due to the minimum tolerance remaining in structural design, the four holes on each part are well aligned for screw threading. The DelTact sensor can be mounted onto the FE gripper of Panda robotic arm directly to perform manipulation tasks (shown in Fig. \ref{fig: cover}.a).


\section{Software Design} \label{sec: software}

In this section, we show how to convert raw input images into meaningful contact information from DelTact. Then rich contact information such as deformation, force distribution, and shape with high resolution and accuracy is extracted online using computationally effective algorithms.
%
\subsection{Image Preprocessing}

As a fisheye lens is used to obtain a large field of view (FOV) and fully cover the sensing area, the camera requires calibration to compensate for the radial and tangential distortion. In addition, distortion also occurs due to the thick silicone layer, which adds a lens effect to the original image. Therefore, we take this into account by calibrating the camera module in the presence of the gel.

The \href{https://docs.opencv.org/3.4/dc/dbb/tutorial_py_calibration.html}{OpenCV} camera calibration functionality is utilized \cite{opencv_library}. To capture the distorted image through silicone, we fabricate a tactile base with a transparent Solaris™ layer (the right one shown in Fig. \ref{fig:tactile base}) and mount it onto the sensor. A chessboard is printed to mark the 3D position of points in the world frame. We mounted the transparent tactile skin to the sensor and took 14 images of the chessboard at different positions and orientations for the intrinsic and extrinsic calibration. 



\subsection{Deformation Measurement with Dense Color Pattern} \label{subsec: Random Color Pattern algo}

Optical flow tracking of the dense random color pattern is the prime algorithm that is utilized to measure sensor surface deformation. The vector field obtained from the algorithm represents the 2D projection of the 3D surface deformation on camera frame, from which rich contact information can be extracted. Thus, tracking accuracy of the pattern influences the sensor performance at a fundamental level. Here we adopt improved optical flow with adaptive referencing, which has been proposed in \cite{du2021high} and is briefly reviewed in Section \ref{subsubsec: dis flow}. Then, we focus on generating color patterns with high randomness (Section \ref{subsubsec: pattern generate}).

\subsubsection{Dense Optical Flow and Adaptive Reference} \label{subsubsec: dis flow}

We pursue a dense optical flow using Gunnar Farneback's algorithm \cite{farneback2003two} on GPU for accuracy and less overhead. The algorithm estimates the 2D displacement vector field from the image sequence at high frequency on GPU \cite{opencv_library}. 
The algorithm solves the traditional optical flow problem in a dense (per pixel) manner, by finding a warping vector \textbf{u} = (\textit{u}, \textit{v}) for each template patch \textit{T} in the reference image which minimizes the squared error between patches in reference image and query image $I_{t}$.

\begin{equation}
    \begin{aligned}
       \text{\textbf{u}} = \text{argmin}_{\text{\textbf{u}}'} \sum_{x}[I_{t}(\text{\textbf{x}} + \text{\textbf{u}}') - T(\text{\textbf{x}})]^{2}
    \end{aligned}
    \label{eq: warping vector}
    \vspace{-0.1cm}
\end{equation}
where \textbf{x} = $(x, y)^T$ represents pixels in patch $T$ from the reference image.

While referring to a static initial frame causes imperfections under large deformation when the rigid template matching fails to track the distorted pattern, an adaptive referencing strategy is introduced to automatically select a new reference frame during operation. An inverse wrapped image is computed based on a coarser flow and compared with the current reference image. Once the photometric error between two images exceeds a fixed threshold, the current image is set to replace the old reference image for tracking. During this process, the total optical flow is the superposition of all the flows calculated. This method guarantees that a matched correspondence between each frame is accurate and allows small non-linear transformation for the template image \cite{du2021high}.

\subsubsection{Pattern Generation} \label{subsubsec: pattern generate}
The purpose of using a dense color pattern is that the dense optical flow algorithm estimates the motion of patches based on the image intensity variance as shown in Eq. \ref{eq: warping vector}. Therefore, an initial frame where every pixel has distinct RGB (or grayscale) values is more random and contains more features to track as opposed to a monochromatic image that is untrackable.

Three parameters, i.e. pattern resolution $h \times w$ (the pattern has a length-width ratio of 1:1), patch size $d$, and randomness regulation factor $r$, are predetermined to form the pattern. Here the patch size determines the length of the square color patch in mm. For instance, as the pattern is printed in an area of $35 \times 35$ mm$^2$, a patch size of 0.15 indicates that each color patch is $0.15 \times 0.15$ mm$^2$. Therefore, the pattern resolution will be $700 \times 700$, and each patch takes up $3 \times 3$ pixels.

To generate the color pattern, we begin with filling the first patch at the top-left of the image. Three numbers are drawn randomly with a uniform distribution from 0 to 1 and applied to RGB channels of the patch. Then the rest of the patches are computed based on the existing patches. A neighbor patch of a particular patch is defined if they are connected through an edge, i.e. forming an 4-neighbor system. Moreover, the randomness regulation factor, $r \in [0,1)$ adjusts the variance of RGB values between neighboring patches. A new patch is filled so that the minimum value of the difference square in each RGB channel between the new patch and its neighbors is larger than $r$. A detailed description of this process is shown in Algorithm \ref{alg:pat_code}.

\begin{algorithm} 
	\caption{Patch Generation} 
	\label{alg:pat_code} 
	\begin{algorithmic}[1]
	
		\REQUIRE ~~\\
		Patch size in pixels, $d$;\\
		Randomness regulation factor between neighbors, $r$ ;\\
		Neighbor patches, $P_1, P_2, .., P_n$;\\
		\ENSURE ~~\\
		Dense color patch, $P$;
		\STATE Initialize zero value matrix $P$ with shape $d \times d \times 3$;
		\FOR{i from 1 to n} 
		\STATE Store RGB values $R_i, G_i, B_i$ of neighbor patch $P_i$;
		\ENDFOR
		\STATE Generate $R, G, B = rand(0,1)$;
		\WHILE{$ min((R-R_i)^2, (G-G_i)^2, (B-B_i)^2) < r$}
		    \STATE Regenerate $R$, $G$, $B$;
		\ENDWHILE
		\STATE Apply $R, G, B$ to RGB channels of $P$;
		\RETURN $P$
	\end{algorithmic} 

\end{algorithm}

We continue to fill the first row and first column of patches, where the neighbors are the left and the upper patches respectively. Finally, all unfilled patches are computed in sequence with row-major order concerning the neighbor patches generated previously. To find the proper parameters of $d$ and $r$ such that the error of optical flow tracking is minimized, an indentation experiment was conducted shown in Section \ref{sec: pat exp}.

\subsection{Tactile Measurement Algorithm}\label{sec: tactile measurement algo}
In this section, we present the algorithm pipeline for extracting tactile information, i.e., shape and contact force, from the image. The experimental results of tactile measurements are presented to demonstrate sensor performance.

\subsubsection{Shape Reconstruction}

The method of 3D shape reconstruction was presented in our previous work \cite{du2021high} based on the optical flow with adaptive referencing mentioned in Section \ref{subsubsec: dis flow}. Because the 2D optical flow is essentially a projection of 3D deformation on camera, an expansion field indicates a deformation in normal direction. Thus, to extract the shift-invariant measurement of expansion, we apply a 2D Gaussian distribution kernel to the flow vectors and accumulate the distribution at each point to obtain the Gaussian density. The covariance matrix is given by:

 \begin{equation}
        \mathbf{Q} = 
        \begin{bmatrix}
        \sigma^2 & 0 \\
        0 & \sigma^2 \\
        \end{bmatrix}.
    \label{eq: covariance}
\end{equation}
    
The relative depth of the surface deformation can be directly estimated from negative Gaussian density. The result of shape reconstruction is shown in Section \ref{sec: shape exp}.

\subsubsection{Contact Force Estimation}

Surface total force (normal force and shear force along x/y-directions) can be inferred from the vector field based on natural Helmholtz-Hodge decomposition (NHHD) \cite{zhang2019effective}. The optical flow $\vec{V}$ is decomposed by
\begin{equation}
        \vec{V} =  \vec{d} + \vec{r} +\vec{h},
        \vspace{-1mm}
\end{equation}
where $\vec{d}$ denotes curl-free component ($\nabla \times \vec{d} = \vec{0}$), $\vec{r}$ denotes divergence-free component ($\nabla \cdot \vec{r} = \vec{0}$), and $\vec{h}$ is harmonic ($\nabla \times \vec{h} = \vec{0}$, $\nabla \cdot \vec{h} = \vec{0}$) \cite{bhatia2014natural}. Then summation of vector norms on $\vec{d}$ and norm of vector summation of $\vec{V}$ can be used to estimate total normal force and shear force.

Sometimes a densely distributed force field is preferred in providing richer information for control purposes. Therefore, we now present a method to break down total force into force distribution. Given the optical flow with NHHD components, $\vec{V} =  \vec{d} + \vec{r} +\vec{h}$, we approximate the normal force and shear forces in x and y directions: $f = \begin{bmatrix} f_{normal} & f_{shearX} & f_{shearY} \end{bmatrix}^T$ at the displacement point $p = (i, j)$ with an linear model:
\begin{equation}
        f= \text{diag}\left ( Ax \right ).
        \vspace{-1mm}
\end{equation}
$A$ is a $3 \times 6$ linear coefficient matrix
\begin{equation}
    A = \begin{bmatrix}
     a_{11} & a_{12} & ... & a_{16} \\
     a_{21} & a_{22} & ... & a_{26} \\
     a_{31} & a_{32} & ... & a_{36}
    \end{bmatrix},
    \vspace{-1mm}
\end{equation}
where $a_{14} = a_{15} = a_{16} = 0$.
And $x$ is a $6 \times 3$ linear term matrix 
\begin{equation}
    x = \begin{bmatrix}
     D_p &  {h_{px}} + {r_{px}}   &  {h_{px}} + {r_{px}} \\[0.2em]
     D_p^2 & \left( {h_{px}} + {r_{px}} \right)^2  & \left( {h_{px}} + {r_{px}} \right)^2  \\[0.2em]
     D_p^3 & \left( {h_{px}} + {r_{px}} \right)^3  & \left( {h_{px}} + {r_{px}} \right)^3 \\[0.2em]
     0 &  {h_{py}} + {r_{py}}   &  {h_{py}} + {r_{py}}  \\[0.2em]
     0 & \left( {h_{py}} + {r_{py}} \right)^2  & \left( {h_{py}} + {r_{py}} \right)^2 \\[0.2em]
     0 & \left( {h_{py}} + {r_{py}} \right)^3  & \left( {h_{py}} + {r_{py}} \right)^3 
    \end{bmatrix},
\end{equation}
where $D_p$ is the processed non-negative Gaussian density, ${r_{px}}$, ${r_{py}}$, ${h_{px}}$ and ${h_{py}}$ are the x and y component of $\vec{r}$ and $\vec{h}$ at point $p$. Then, $A$ is calibrated by the total force which is assumed to be the superposition of $f$ across the surface. The total forces along normal and shear directions $F= \begin{bmatrix} F_{normal} & F_{shearX} & F_{shearY} \end{bmatrix}^T$ are also defined by the linear model as:

\begin{equation}
        F = \text{diag}\left ( AX \right ),
\end{equation}
with  
\begin{equation}
    X = \begin{bmatrix}
     \sum D_p &  \sum \left( {h_{px}} + {r_{px}} \right)  &  \sum \left( {h_{px}} + {r_{px}} \right) \\[0.2em]
     \sum D_p^2 & \sum \left( {h_{px}} + {r_{px}} \right)^2  & \sum \left( {h_{px}} + {r_{px}} \right)^2  \\[0.2em]
     \sum D_p^3 & \sum \left( {h_{px}} + {r_{px}} \right)^3  & \sum {\left( {h_{px}} + {r_{px}} \right)^3} \\[0.2em]
     0 &  \sum \left( {h_{py}} + {r_{py}} \right)  &  \sum \left( {h_{py}} + {r_{py}} \right)  \\[0.2em]
     0 & \sum \left( {h_{py}} + {r_{py}} \right)^2  & \sum \left( {h_{py}} + {r_{py}} \right)^2 \\[0.2em]
     0 & \sum \left( {h_{py}} + {r_{py}} \right)^3  & \sum \left( {h_{py}} + {r_{py}} \right)^3 
    \end{bmatrix}.
\end{equation}

This model is calibrated with another force indentation experiment as shown in Section \ref{sec: force cali}.

\section{Experiments and Results}\label{sec: experiment}
In this section, we present the experiment steps and results of random pattern selection and force regression model calibration with two similar indentation experiments. Then the experimental results of tactile measurements, i.e., to extract shape and contact force from the optical flow, are presented to demonstrate sensor performance.

\subsection{Pattern Selection} \label{sec: pat exp}

\begin{figure} 
    \centering
    \vspace*{3mm}

    \includegraphics[width=0.475\textwidth]{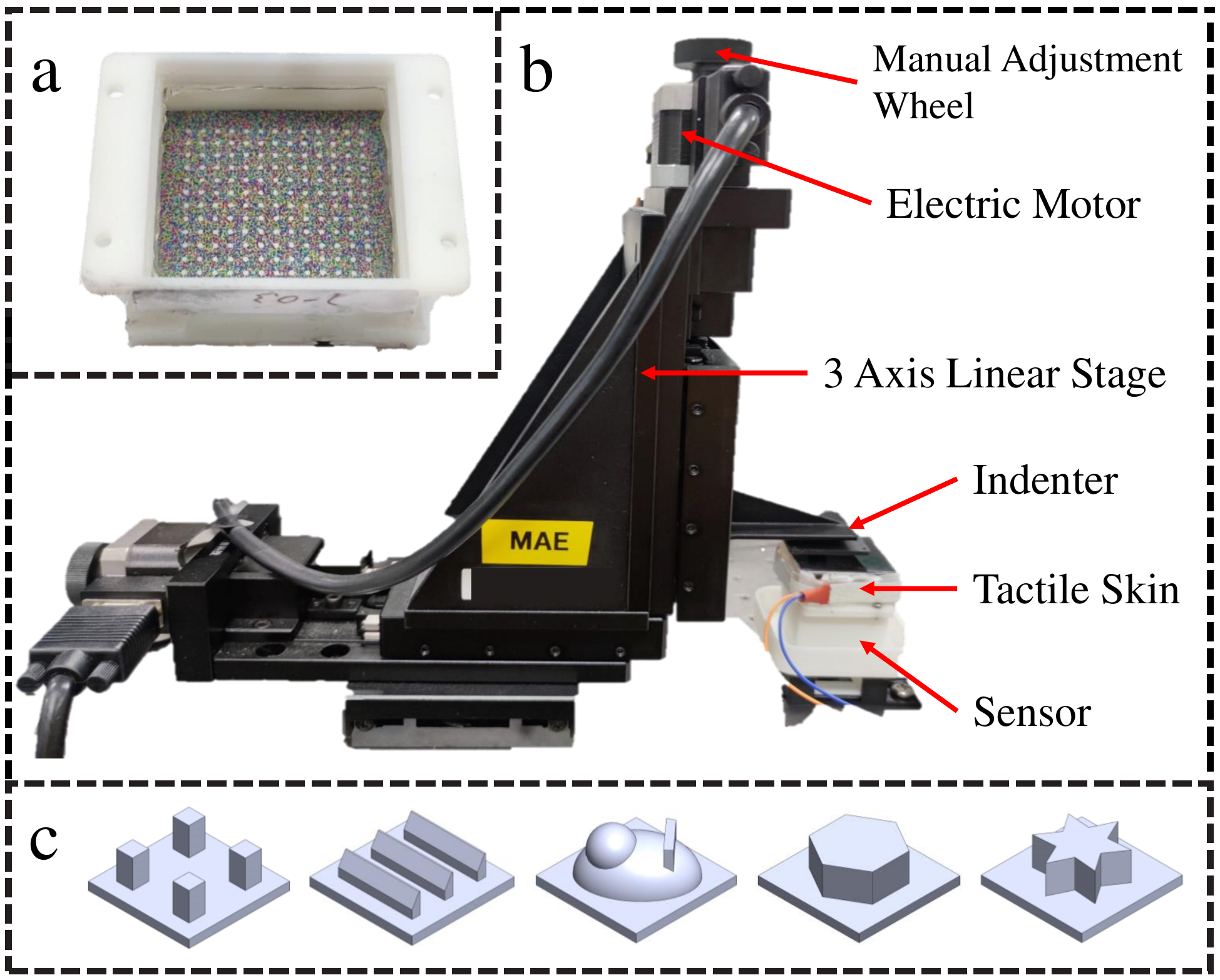}
    \vspace*{-3mm}
    \caption{Pattern selection experiment configuration. a) Demountable tactile skin base using dense color combined with white dots pattern($d=0.2$, $r=0.3$). b) Data collection with electric 3 axis linear stage. c) Solidworks modelings of 5 indenters. From left to right: 4 dots, edges, ellipsoid, hexagonal prism, star.}
    \label{fig:pat exp config}
    \vspace{-7mm}
\end{figure}

Because the ink printing quality and the camera resolution are limited, there are a minimal patch size and maximal randomness for the pattern to be captured by the camera.
To choose the proper dense color pattern that fits well with the optical flow method, we conducted a series of indentation experiments to measure the accuracy of flow tracking using different patterns. We first fabricated nine tactile skin bases with patch size $d \in \{ 0.1, 0.15, 0.2 \}$ and randomness regulation factor $r \in \{ 0.1, 0.3, 0.5 \} $. The skin bases were mounted on a testing sensor that had the same configuration as DelTact but a different shell for fixing on a table (shown in Fig. \ref{fig:pat exp config}.b). Five 3D printed indenters (shown in Fig. \ref{fig:pat exp config}.c) pressed the sensor surface and moved along x/y/z-axis by the electric linear stage to generate surface deformation in all directions. 

The experiment was carried out in the following steps:

\begin{enumerate}
  \item The sensor with tactile base mounted was fixed on a table. 
  \item The 2-dot indenter was installed on the linear stage.
  \item The stage was driven to press the sensor surface at four positions, the contact depths at each position were 5 mm and then 10 mm. 
  \item At each depth, the indenter moved in x/y-directions for $\pm 10$ mm. During this process, the camera continued to capture the image.
  \item The stage retracted to its initial position to change the indenter. Then step 3 to 5 were repeated.
  \item After five indenters were used, the tactile base was replaced, and data collection was repeated from step 2.
\end{enumerate}

For each combination of indenter and tactile base, 3600 data points were collected. Each data point was an image with deformed dense color pattern with dot markers embedded inside (a sample is shown in Fig. \ref{fig:pat exp config}.a). A color filter separated the white color dots in the image and a blob detection measured the sub-pixel displacement vectors, $\textbf{u}_i = (dx_i, dy_i)$, from current position to initial position at $p_i = (x_i,y_i)$ ($i$ = 1,2,3...169). As the tracking from blob detection could reach sub-pixel accuracy, the white dot displacements were regarded as the ground truth to compare with optical flow at corresponding positions. We then ran the dense optical algorithm and calculated the average error $\delta$ between the flow displacement vectors $\textbf{u}'_i = (dx'_i, dy'_i)$ at $p_i$ and $\textbf{u}_i$ at $p_i$. For $n = 169$, the error $\Bar{\delta}$ was given by:

\begin{equation}
    \begin{aligned}
        \Bar{\delta}=\frac{\sum_{i=1}^{n}\sqrt{\left(\left(x_i-x'_i\right)^{2}+\left(y_i-y'_i\right)^{2}\right)}}{n}
    \end{aligned}
    \label{eq: tracking error}
\end{equation}

\begin{figure} 
    \centering
    \includegraphics[width=0.478\textwidth]{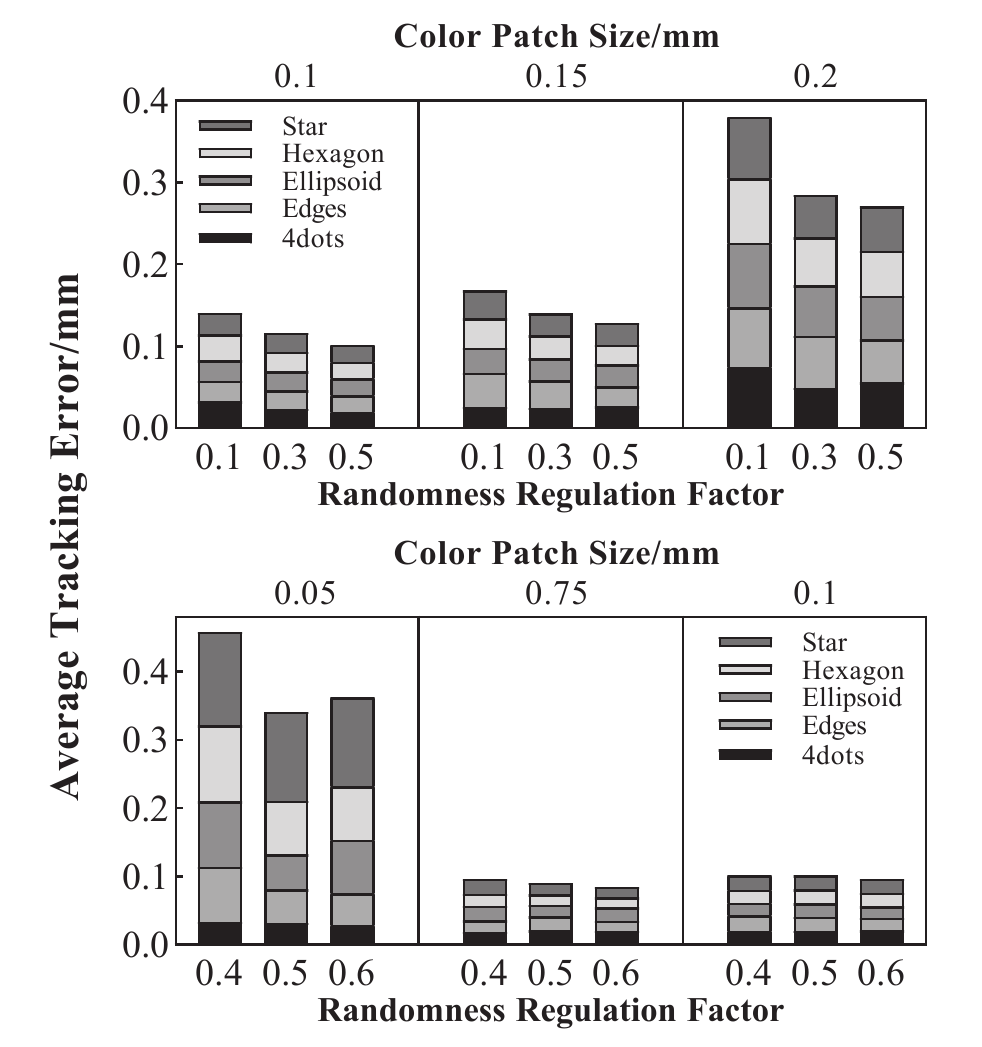}
    \vspace*{-0.4cm}
    \caption{Compression of the average optical flow tracking error between patterns with different patch size $d$ and random regulation factor $r$.}
    \label{fig:pat exp result}
    \vspace{-0.2cm}
\end{figure}

The experiment results are shown in the upper one of the stacked column charts in Fig. \ref{fig:pat exp result}. The average tracking errors for each pattern under five indenters are accumulated to evaluate the performance under different contact shapes. The pattern with $d = 0.1$ and $r = 0.5$ gives the lowest error of 0.11 mm. And we continued to conduct another indentation to obtain a sub-optimal result with $r$ and $d$ near this value, given $d \in \{ 0.5, 0.75, 0.1 \}$ and $r \in \{ 0.4, 0.5, 0.6 \}$. As shown in the lower chart in Fig. \ref{fig:pat exp result}, when $d = 0.075$ and $r = 0.6$, the error reaches the lowest of 0.08 mm. Therefore, this pattern is used to fabricate our sensor. We can see that patch size $d$ dominates the error. When $d$ is small enough, increasing $r$ does not have significant influence on result. It is also noticed that as the $d$ becomes extremely smaller ($d = 0.05$), the error increases (around 0.5 mm) instead of drop. The reason is that the color patch is such small that is beyond the maximum resolution that the printer can fabricate. This results in a ambiguous gray area in the pattern causing mis-tracking in the optical flow algorithm. But in the rest of the experimental range, the tracking error decreases when the patch size decreases and randomness regulation factor increases, which agrees with our initial purpose of using a more random and denser color pattern to obtain more accurate displacement tracking. 

\subsection{Shape Reconstruction}\label{sec: shape exp}

\begin{figure}
    \centering
    \includegraphics[width=0.48\textwidth]{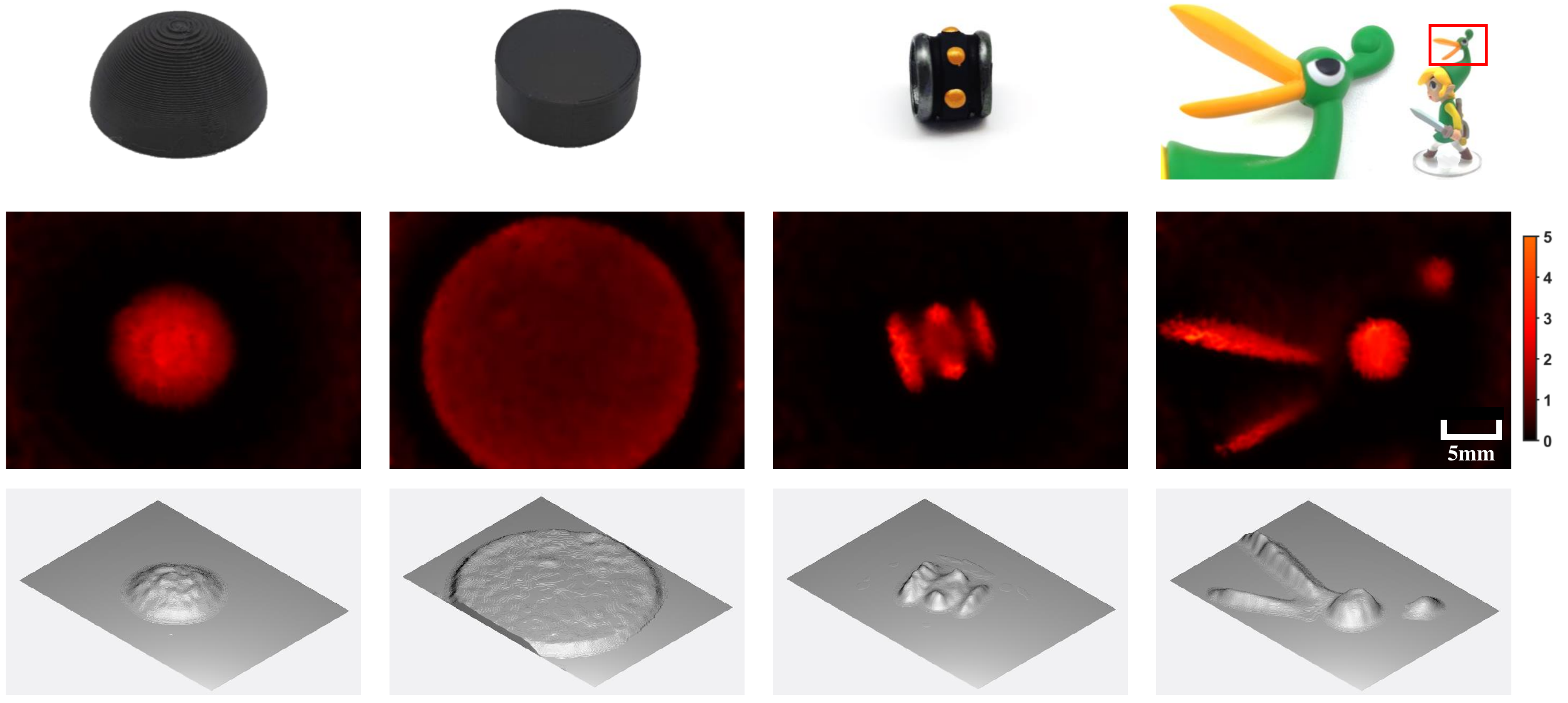}
    \vspace*{-4mm}
    \caption{3D reconstruction of four shapes (from left to right): sphere, cylinder, ring, and toy hat. The first row are the objects. The second row is the estimated Gaussian densities plotted in hot map. A more reddish region indicates a higher depth. The third row is the isometric view of the deformation.}
    \label{fig:3d rec}
    \vspace{-0.6cm}
\end{figure}

In practice, we set $\sigma=3.0$, and a guided filter \cite{he2015fast} was used to reduce high-frequency noises and smooth the surface while maintaining shape features such as edges. We conducted the shape reconstruction for four objects in Fig. \ref{fig:3d rec}. The 3D shape of the contact objects is shown in the results, including the features of faces, edges, curves, and corners.

\subsection{Contact Force Distribution}\label{sec: force cali}

To calibrate the contact force model, we collected force and flow data by conducting a similar indentation experiment as shown in Section \ref{sec: pat exp}. An ATI Nano17 F/T Sensor was installed on the sensor to measure high-accuracy surface normal and shear force. As for the indenters, we used five 3D printed spheres with diameters of 10, 12, 15, 18, and 22 mm. Each indenter pressed the sensor at 9 positions and moved 5 normal steps and 9 shear steps to load different normal and shear forces. To avoid the influence of slip, only steady-state data were recorded. $A$ was solved using linear regression with $ 9 \times \left( 5 \times 9 + 1 \right) \times 5 = 2070$ data points, where 1656 data points were used for training and 414 for testing. The range of measured shear force during experiment was from -2.49 N to 2.94 N in x-direction, and -2.82 N to 2.86 N in y-direction. The range of normal force was from 0 N to 9.67 N. We adopted $n = 3$, and the resulting adjusted coefficient of determination $R^2$ together with root mean square error (RMSE) are shown in Table. \ref{Tab:linearR}.

\begin{table}
\vspace*{2mm}
\caption{Linear Regression Analysis}
\vspace*{-3mm}
\centering
    \begin{tabular}{l|cc}
    \textbf{Force} & Adjusted $R^2$ & RMSE (N) \\ \hline
    $F_{normal}$     & 0.99  & 0.30    \\
    $F_{shearX}$     & 0.98  & 0.14   \\
    $F_{shearY}$     & 0.98  & 0.17  
    \end{tabular}
    \label{Tab:linearR}
\vspace*{-6mm}
\end{table}

\begin{figure}  [ht]
    \centering
    \includegraphics[width=0.48\textwidth]{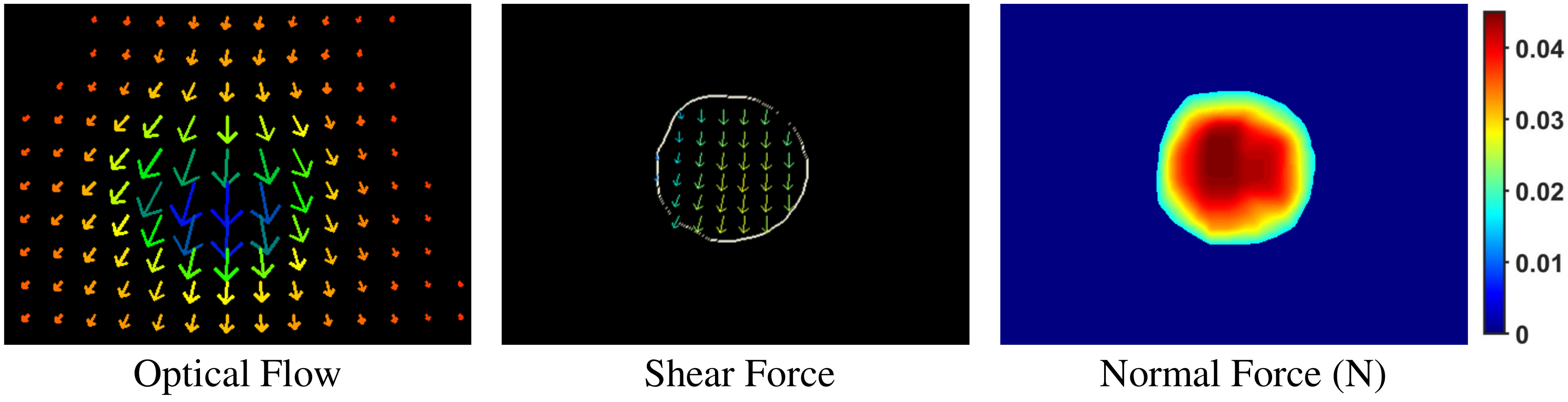}
    \vspace*{-7mm}
    \caption{Force distribution estimation result of a sphere indenter. For visualization, the dense vector field of optical flow and shear force are sparsely displayed. A white contour representing the contact area is shown in the shear force distribution.}
    \label{fig: force dis}
\end{figure}

The RMSE is 0.3 N, 0.14 N and 0.17 N for normal and shear force in x/y-directions. Considering that this is the error of surface total force, the actual error of force estimation will be lower if divided at each point on the surface. 
Finally, the calibrated model was applied to the sensor, where we revealed the force distribution results in Fig. \ref{fig: force dis}. The algorithm achieved an online computation frequency of 40 Hz with an Intel Core i7-7700 CPU and an NVIDIA GTX 1060 GPU. 

\section{Discussion and Conclusion}\label{sec: conclusion}

In this work, we present the design of a new vision-based tactile sensor with an optimized dense random color pattern. The proposed sensor, named DelTact, adopts reinforced hardware that features greater compactness and robustness. It can be mounted onto various types of end-effectors with redesigned sensor connection part. The size of DelTact is reduced by one-third compared to the fingervision sensor \cite{du2021high} whilst keeping the sensory area sufficient for contact measurement. Random color patterns generated from different parameter sets were tested with an indentation experiment for minimal tracking error. Regarding software, image preprocessing, shape reconstruction, and contact force estimation algorithms are presented with experimental results showcasing that our sensor has multi-modality sensing abilities with high resolution and frequency. A comparison between DelTact and other vision-based tactile sensors is presented in Table. \ref{Tab:sensor comp}. From the table, we can see that our sensor provides a large sensing area at a higher resolution with a compact size.



\begin{table}[ht]
\scriptsize
\vspace*{3mm}
\caption{Sensor Comparison}
\vspace*{-2mm}
\centering
    \begin{tabular}{l|c|>{\centering\arraybackslash}m{1cm}|>{\centering\arraybackslash}m{1cm}|c|>{\centering\arraybackslash}m{1cm}}
\hline 
Sensor     & Resolution & Sensing Area (mm$^2$)& Pixel Size (mm) & Size (mm$^3$) & Calibration\\ \hline 
\cite{dong2017improved} & \multirow{6}{*}{640x480} & 252 & 0.028 & 40x80x40 &  $\checkmark$ \\ 
\cite{taylor2021gelslim3} &  &       675  & 0.047  & 37x80x20 &  $\checkmark$ \\ 
\cite{lambeta2020digit}   &  &       305  & 0.031   & 20x27x18 &  $\times$\\ 
\cite{yamaguchi2016combining} &  & 750 &0.049& 40x47x30 & $\checkmark$ \\ 
\cite{sferrazza2019design}  &   &    900  &0.054&  50x50x43.8 & $\checkmark$ \\
\cite{du2021high} &  & 756 & 0.049& 20x20x26 & $\times$ \\ \hline
\cite{ward2018tactip} & 162  & 628&1.96 & 20x20x26 & $\times$ \\ \hline

\textbf{Ours}   & 798x586 &   675 &0.037 & 39x60x30 & $\checkmark$ \\ \hline 
    \end{tabular}
    \label{Tab:sensor comp}
\vspace*{-6mm}
\end{table}

We acknowledge the compromise between the information quality and limitation of sensor performance, while rough contact feedback can satisfy the system perception demand in some cases with lower requirements in hardware and software. The proposed methods of shape reconstruction with Gaussian density falls short compared with prior works in texture measurement based on photometric stereo such as GelSight \cite{dong2017improved} and GelSlim \cite{taylor2021gelslim3}, which are able to recognize surface features at sub-millimeter scale. However, photometric stereo requires strict conditions for surface reflection and illumination properties. Learning-based force estimation manages to measure contact force within an error of 0.1 N \cite{sferrazza2019design}, but the confidence of the model prediction relies on a large amount of training data (over 10000) from long collecting procedures. Therefore, the motivation of our work is to devise an easily fabricated and calibrated sensor that is sufficient and cost-effective in tactile information extraction for a broader range of tasks.

Possible future work includes testing the versatility of the sensor on robot perception and manipulation. Beyond the low-level features, using optical flow, we may try to extract higher-level features such as vibration and slip, which are critical for maintaining stability in grasping tasks. Besides, we aim to obtain higher accuracy 3D point cloud of surface deformation to remove noise in shape reconstruction .


\bibliographystyle{IEEEtran}
\bibliography{reference}

\begin{thebibliography}{10}
\providecommand{\url}[1]{#1}
\csname url@samestyle\endcsname
\providecommand{\newblock}{\relax}
\providecommand{\bibinfo}[2]{#2}
\providecommand{\BIBentrySTDinterwordspacing}{\spaceskip=0pt\relax}
\providecommand{\BIBentryALTinterwordstretchfactor}{4}
\providecommand{\BIBentryALTinterwordspacing}{\spaceskip=\fontdimen2\font plus
\BIBentryALTinterwordstretchfactor\fontdimen3\font minus
  \fontdimen4\font\relax}
\providecommand{\BIBforeignlanguage}[2]{{%
\expandafter\ifx\csname l@#1\endcsname\relax
\typeout{** WARNING: IEEEtran.bst: No hyphenation pattern has been}%
\typeout{** loaded for the language `#1'. Using the pattern for}%
\typeout{** the default language instead.}%
\else
\language=\csname l@#1\endcsname
\fi
#2}}
\providecommand{\BIBdecl}{\relax}
\BIBdecl

\bibitem{zou2017novel}
L.~Zou, C.~Ge, Z.~J. Wang, E.~Cretu, and X.~Li, ``Novel tactile sensor
  technology and smart tactile sensing systems: A review,'' \emph{Sensors},
  vol.~17, no.~11, p. 2653, 2017.

\bibitem{yuan2017gelsight}
W.~Yuan, S.~Dong, and E.~H. Adelson, ``Gelsight: High-resolution robot tactile
  sensors for estimating geometry and force,'' \emph{Sensors}, vol.~17, no.~12,
  p. 2762, 2017.

\bibitem{taylor2021gelslim3}
I.~Taylor, S.~Dong, and A.~Rodriguez, ``Gelslim3. 0: High-resolution
  measurement of shape, force and slip in a compact tactile-sensing finger,''
  \emph{arXiv preprint arXiv:2103.12269}, 2021.

\bibitem{lambeta2020digit}
M.~Lambeta, P.-W. Chou, S.~Tian, B.~Yang, B.~Maloon, V.~R. Most, D.~Stroud,
  R.~Santos, A.~Byagowi, G.~Kammerer \emph{et~al.}, ``Digit: A novel design for
  a low-cost compact high-resolution tactile sensor with application to in-hand
  manipulation,'' \emph{IEEE Robotics and Automation Letters}, vol.~5, no.~3,
  pp. 3838--3845, 2020.

\bibitem{vlack2004gelforce}
K.~Vlack, K.~Kamiyama, T.~Mizota, H.~Kajimoto, N.~Kawakami, and S.~Tachi,
  ``Gelforce: A traction field tactile sensor for rich human-computer
  interaction,'' in \emph{IEEE Conference on Robotics and Automation, 2004.
  TExCRA Technical Exhibition Based.}\hskip 1em plus 0.5em minus 0.4em\relax
  IEEE, 2004, pp. 11--12.

\bibitem{sferrazza2019design}
C.~Sferrazza and R.~D’Andrea, ``Design, motivation and evaluation of a
  full-resolution optical tactile sensor,'' \emph{Sensors}, vol.~19, no.~4, p.
  928, 2019.

\bibitem{kuppuswamy2020soft}
N.~Kuppuswamy, A.~Alspach, A.~Uttamchandani, S.~Creasey, T.~Ikeda, and
  R.~Tedrake, ``Soft-bubble grippers for robust and perceptive manipulation,''
  in \emph{2020 IEEE/RSJ International Conference on Intelligent Robots and
  Systems (IROS)}.\hskip 1em plus 0.5em minus 0.4em\relax IEEE, 2020, pp.
  9917--9924.

\bibitem{du2021high}
Y.~Du, G.~Zhang, Y.~Zhang, and M.~Y. Wang, ``High-resolution 3-dimensional
  contact deformation tracking for fingervision sensor with dense random color
  pattern,'' \emph{IEEE Robotics and Automation Letters}, vol.~6, no.~2, pp.
  2147--2154, 2021.

\bibitem{kamiyama2001vision}
K.~Kamiyama, H.~Kajimoto, M.~Inami, N.~Kawakami, and S.~Tachi, ``A vision-based
  tactile sensor,'' in \emph{International Conference on Artificial Reality and
  Telexistence}, 2001, pp. 127--134.

\bibitem{yamaguchi2016combining}
A.~Yamaguchi and C.~G. Atkeson, ``Combining finger vision and optical tactile
  sensing: Reducing and handling errors while cutting vegetables,'' in
  \emph{2016 IEEE-RAS 16th International Conference on Humanoid Robots
  (Humanoids)}.\hskip 1em plus 0.5em minus 0.4em\relax IEEE, 2016, pp.
  1045--1051.

\bibitem{ward2018tactip}
B.~Ward-Cherrier, N.~Pestell, L.~Cramphorn, B.~Winstone, M.~E. Giannaccini,
  J.~Rossiter, and N.~F. Lepora, ``The tactip family: Soft optical tactile
  sensors with 3d-printed biomimetic morphologies,'' \emph{Soft robotics},
  vol.~5, no.~2, pp. 216--227, 2018.

\bibitem{dong2017improved}
S.~Dong, W.~Yuan, and E.~H. Adelson, ``Improved gelsight tactile sensor for
  measuring geometry and slip,'' in \emph{2017 IEEE/RSJ International
  Conference on Intelligent Robots and Systems (IROS)}.\hskip 1em plus 0.5em
  minus 0.4em\relax IEEE, 2017, pp. 137--144.

\bibitem{padmanabha2020omnitact}
A.~Padmanabha, F.~Ebert, S.~Tian, R.~Calandra, C.~Finn, and S.~Levine,
  ``Omnitact: A multi-directional high-resolution touch sensor,'' in \emph{2020
  IEEE International Conference on Robotics and Automation (ICRA)}.\hskip 1em
  plus 0.5em minus 0.4em\relax IEEE, 2020, pp. 618--624.

\bibitem{li2020review}
Q.~Li, O.~Kroemer, Z.~Su, F.~F. Veiga, M.~Kaboli, and H.~J. Ritter, ``A review
  of tactile information: Perception and action through touch,'' \emph{IEEE
  Transactions on Robotics}, vol.~36, no.~6, pp. 1619--1634, 2020.

\bibitem{donlon2018gelslim}
E.~Donlon, S.~Dong, M.~Liu, J.~Li, E.~Adelson, and A.~Rodriguez, ``Gelslim: A
  high-resolution, compact, robust, and calibrated tactile-sensing finger,'' in
  \emph{2018 IEEE/RSJ International Conference on Intelligent Robots and
  Systems (IROS)}.\hskip 1em plus 0.5em minus 0.4em\relax IEEE, 2018, pp.
  1927--1934.

\bibitem{she2019cable}
Y.~She, S.~Wang, S.~Dong, N.~Sunil, A.~Rodriguez, and E.~Adelson, ``Cable
  manipulation with a tactile-reactive gripper,'' \emph{arXiv preprint
  arXiv:1910.02860}, 2019.

\bibitem{yuan2015measurement}
W.~Yuan, R.~Li, M.~A. Srinivasan, and E.~H. Adelson, ``Measurement of shear and
  slip with a gelsight tactile sensor,'' in \emph{2015 IEEE International
  Conference on Robotics and Automation (ICRA)}.\hskip 1em plus 0.5em minus
  0.4em\relax IEEE, 2015, pp. 304--311.

\bibitem{cramphorn2018voronoi}
L.~Cramphorn, J.~Lloyd, and N.~F. Lepora, ``Voronoi features for tactile
  sensing: Direct inference of pressure, shear, and contact locations,'' in
  \emph{2018 IEEE International Conference on Robotics and Automation
  (ICRA)}.\hskip 1em plus 0.5em minus 0.4em\relax IEEE, 2018, pp. 2752--2757.

\bibitem{yuan2017connecting}
W.~Yuan, S.~Wang, S.~Dong, and E.~Adelson, ``Connecting look and feel:
  Associating the visual and tactile properties of physical materials,'' in
  \emph{Proceedings of the IEEE Conference on Computer Vision and Pattern
  Recognition}, 2017, pp. 5580--5588.

\bibitem{li2014localization}
R.~Li, R.~Platt, W.~Yuan, A.~ten Pas, N.~Roscup, M.~A. Srinivasan, and
  E.~Adelson, ``Localization and manipulation of small parts using gelsight
  tactile sensing,'' in \emph{2014 IEEE/RSJ International Conference on
  Intelligent Robots and Systems (IROS)}.\hskip 1em plus 0.5em minus
  0.4em\relax IEEE, 2014, pp. 3988--3993.

\bibitem{bauza2019tactile}
M.~Bauza, O.~Canal, and A.~Rodriguez, ``Tactile mapping and localization from
  high-resolution tactile imprints,'' in \emph{2019 International Conference on
  Robotics and Automation (ICRA)}.\hskip 1em plus 0.5em minus 0.4em\relax IEEE,
  2019, pp. 3811--3817.

\bibitem{pang2021viko}
C.~Pang, K.~Mak, Y.~Zhang, Y.~Yang, Y.~A. Tse, and M.~Y. Wang, ``Viko: An
  adaptive gecko gripper with vision-based tactile sensor,'' in \emph{2021 IEEE
  International Conference on Robotics and Automation (ICRA)}, 2021, pp.
  736--742.

\bibitem{opencv_library}
G.~Bradski, ``{The OpenCV Library},'' \emph{Dr. Dobb's Journal of Software
  Tools}, 2000.

\bibitem{farneback2003two}
G.~Farneb{\"a}ck, ``Two-frame motion estimation based on polynomial
  expansion,'' in \emph{Scandinavian conference on Image analysis}.\hskip 1em
  plus 0.5em minus 0.4em\relax Springer, 2003, pp. 363--370.

\bibitem{zhang2019effective}
Y.~Zhang, Z.~Kan, Y.~Yang, Y.~A. Tse, and M.~Y. Wang, ``Effective estimation of
  contact force and torque for vision-based tactile sensors with
  helmholtz–hodge decomposition,'' \emph{IEEE Robotics and Automation
  Letters}, vol.~4, no.~4, pp. 4094--4101, 2019.

\bibitem{bhatia2014natural}
H.~Bhatia, V.~Pascucci, and P.-T. Bremer, ``The natural helmholtz-hodge
  decomposition for open-boundary flow analysis,'' \emph{IEEE transactions on
  visualization and computer graphics}, vol.~20, no.~11, pp. 1566--1578, 2014.

\bibitem{he2015fast}
K.~He and J.~Sun, ``Fast guided filter,'' \emph{arXiv preprint
  arXiv:1505.00996}, 2015.

\end{thebibliography}

\end{document}